\documentclass[letterpaper]{article} 
\usepackage{aaai2026}  
\usepackage{amssymb}
\usepackage{times}  
\usepackage{helvet}  
\usepackage{courier}  

\usepackage{marvosym} 

\usepackage[hyphens]{url}  
\usepackage{graphicx} 
\usepackage{natbib}  
\usepackage{caption} 
\usepackage{algorithm}
\usepackage{algorithmic}
\usepackage{svg}
\usepackage{times}
\usepackage{latexsym}
\usepackage{amsmath}
\usepackage{graphicx}
\usepackage{amssymb}
\usepackage{booktabs}
\usepackage{makecell}

\usepackage{pifont}
\usepackage{soul}
\usepackage{marvosym}  

\usepackage[T1]{fontenc}
\usepackage[utf8]{inputenc}

\usepackage{microtype}
\usepackage{inconsolata}

\usepackage{pifont}
\usepackage{marvosym} 

\usepackage{newfloat}
\usepackage{listings}
\DeclareCaptionStyle{ruled}{labelfont=normalfont,labelsep=colon,strut=off} 
\floatstyle{ruled}
\newfloat{listing}{tb}{lst}{}
\floatname{listing}{Listing}

\title{How to make Medical AI Systems safer? Simulating Vulnerabilities, \\and Threats in Multimodal Medical RAG Systems}

\author{
  Kaiwen Zuo\textsuperscript{1}\thanks{Main contribution.},
  Zelin Liu\textsuperscript{2},
  Raman Dutt\textsuperscript{3},
  Ziyang Wang\textsuperscript{4},
  Zhongtian Sun\textsuperscript{5},
  Fan Mo\textsuperscript{6}\thanks{\Letter~Corresponding author.},
  Pietro Li\`o\textsuperscript{7}\footnotemark[2]
}

\affiliations{
    \textsuperscript{\rm 1}University of Warwick, Kaiwen.Zuo@warwick.ac.uk;
    \textsuperscript{\rm 2}Shanghai Jiao Tong University, julio@stju.edu.cn;
    \textsuperscript{\rm 3}University of Edinburgh, raman.dutt@ed.ac.uk;
    \textsuperscript{\rm 4}Aston University, z.wang47@aston.ac.uk;
    \textsuperscript{\rm 5}University of Kent, zs256@kent.ac.uk;
    \textsuperscript{\rm 6}University of Oxford, kell6420@ox.ac.uk;
    \textsuperscript{\rm 7}University of Cambridge, pl219@cam.ac.uk
}

\usepackage{bibentry}

\begin{document}

\maketitle

\begin{abstract}
Large Vision–Language Models (LVLMs) augmented with Retrieval-Augmented Generation (RAG) are increasingly used in medical AI for improved factual grounding via external clinical image-text retrieval. However, this reliance introduces a significant attack surface. We propose \textit{MedThreatRAG}, a novel multimodal poisoning framework that systematically probes vulnerabilities in medical RAG systems by injecting adversarial image-text pairs. A key innovation of our approach is the construction of a \textbf{simulated semi-open attack environment}, mimicking real-world medical systems that permit periodic knowledge base updates through user or pipeline contributions. Within this setting, we introduce and emphasize \textbf{Cross-Modal Conflict Injection} (CMCI), which embeds subtle semantic contradictions between medical images and their paired reports. These mismatches degrade retrieval and generation by disrupting cross-modal alignment, while remaining plausible enough to evade conventional filters. While basic textual and visual attacks are also included for completeness, CMCI demonstrates the most severe degradation. Evaluations on IU-Xray and MIMIC-CXR QA show that MedThreatRAG reduces answer F1 scores by up to \textbf{27.66\%} and lowers LLaVA-Med-1.5 F1 rates to as low as \textbf{51.36\%}. Our findings expose fundamental security gaps in clinical RAG systems and highlight the urgent need for threat-aware design and robust multimodal consistency checks. Finally, we also conclude with a concise set of guidelines to inform the safe development of future multimodal medical RAG systems.
\end{abstract}


\section{Introduction}

Medical Vision-Language Models (Med-VLMs), which integrate visual and textual modalities, are becoming increasingly central to numerous clinical tasks, including radiology report generation, Visual Question Answering (VQA), and diagnostic decision support~\cite{yuan2023ramm, hartsock2024vision, chen2024survey}. By jointly encoding medical images (e.g., X-rays, MRIs) and clinical text (e.g., questions, reports), these models offer a unified framework for multimodal understanding in healthcare.

Despite their promise, the trustworthiness of Med-VLMs remains a critical concern. Hallucinations, confident but clinically incorrect outputs, pose a serious threat in medical applications~\cite{zhou2023survey}, where even a minor factual error in generated content can lead to severe or even life-threatening consequences. Improving factual consistency and grounding remains an open challenge in clinical AI systems~\cite{miura2023factual}.

To address this issue, RAG has emerged as a widely adopted solution. By retrieving relevant external knowledge, such as image-report pairs or medical literature~\cite{jain2021knowledge}, RAG augments the generation process with domain-specific evidence, effectively reducing hallucinations and improving factual correctness~\cite{zhao2024retrieval, wu2024medical}. Recent work, such as~\cite{xia-etal-2024-rule}, has demonstrated that RAG significantly boosts both the accuracy and reliability of medical large vision-language models when properly adapted to the domain.

However, while RAG reduces hallucinations, it also introduces a new and underexplored vulnerability: sensitivity to data poisoning. In real-world deployments, external knowledge bases often undergo periodic updates, including user contributions or automated data pipelines~\cite{jafarzadeh2025evidence}. This semi-open structure opens the door to adversaries who can inject harmful, misleading, or semantically conflicting image-text pairs~\cite{carlini2023poisoning} into the retrieval corpus, without needing access to model weights or prompts.

Ensuring the security of medical AI models is urgent~\cite{finlayson2019adversarial} - contamination of external medical knowledge can severely compromise their credibility and clinical value. Thus, safeguarding RAG-based systems from retrieval-stage manipulation is not just a technical issue but a clinical imperative.

To this end, we introduce \textbf{MedThreatRAG}, a novel adversarial attack framework that systematically targets RAG pipelines in Med-VLMs. Rather than manipulating model inputs directly, our approach corrupts the intermediate retrieval process, injecting adversarial image-text pairs that are semantically plausible yet clinically misleading. Our main contributions are as follows:
\begin{itemize}
    \item \textbf{MedThreatRAG} is proposed as a systematic attack framework comprising three complementary strategies—\textit{Textual Attack}, \textit{Visual Attack}, and \textit{Cross-Modal Conflict Injection}—designed to compromise different stages of the RAG pipeline, including the retriever, reranker, and generator.

    \item We construct a \textbf{simulated semi-open attack environment} that mirrors real-world deployment conditions in clinical AI systems, where external knowledge bases are updated via user contributions or automated pipelines. This setting enables attackers to inject Malicious multimodal content without requiring access to model weights or internal parameters.

    \item Extensive evaluations are conducted on IU-Xray and MIMIC-CXR, showing that the proposed framework leads to substantial drops in precision, recall, and F1 scores across state-of-the-art Med-VLMs, while preserving high semantic plausibility.
\end{itemize}

The remainder of this paper is organized as follows: Section~\ref{sec:Preliminaries} describes the multimodal RAG pipeline and generative diffusion process. Section~\ref{sec:Methodology} details the MedThreatRAG architecture and attack strategies. Section~\ref{sec:Experiment} presents experimental results and analyses. Section~\ref{sec:RelatedWork} reviews related literature. Section~\ref{sec:conclusion} concludes the paper and discusses future directions.
\section{Preliminaries}
\label{sec:Preliminaries}
\subsection{Medical Multimodal RAG}
Medical Multimodal RAG enhances parametric knowledge by retrieving relevant medical texts and clinical images from an external Knowledge Base (KB)~\cite{chen2024mllmstrongrerankeradvancing}, aiding more accurate and grounded generation in the biomedical domain. Inspired by recent advances in multimodal RAG~\cite{zhao2023retrieving}, we develop a medical multimodal RAG pipeline consisting of a multimodal medical KB, a domain-specific retriever, a reranker, and a generator. 

Given an input query $q^{(i)}$, the retriever selects the top-$N$ relevant clinical image-report pairs $\{ (x_1, r_1), \cdots, (x_N, r_N) \}$, where $x_j$ represents a medical image (e.g., X-ray, MRI) and $r_j$ is its corresponding radiology report. A CLIP-based retriever, trained on biomedical data, encodes both modalities into a joint space, retrieving candidates based on cosine similarity between the query embedding and each image-text pair.

\subsection{Diffusion Process for Generative Modeling}

In the context of medical image analysis, diffusion models have to be first adapted (fine-tuned) to the medical domain for generating accurate synthetic radiographs ~\cite{bluethgen2024vision,dutt2024parameterefficient} . Diffusion probabilistic models employ a two-phase procedure: a forward diffusion process and a reverse denoising process~\cite{zhang2025frdiff,yang2023diffusion}. In the forward process, a data sample $x_0$ drawn from the true data distribution $q(x_0)$ is gradually corrupted over $T$ time steps by adding Gaussian noise. Each step follows a predefined noise schedule $\beta_t$, resulting in the following Markov process:

\begin{equation}
q(x_t \mid x_{t-1}) = \mathcal{N}(x_t; \sqrt{1 - \beta_t} \, x_{t-1}, \beta_t \mathbf{I}),
\label{eq:forward-step}
\end{equation}
where $\mathbf{I}$ denotes the identity matrix, representing independent and isotropic Gaussian noise added to each dimension of the data.
The marginal distribution of $x_t$ given $x_0$ has a closed-form expression:

\begin{equation}
x_t = \sqrt{\bar{\alpha}_t} \, x_0 + \sqrt{1 - \bar{\alpha}_t} \, \epsilon,
\label{eq:forward-marginal}
\end{equation}

where $\epsilon \sim \mathcal{N}(0, \mathbf{I})$, and $\bar{\alpha}_t = \prod_{i=1}^{t}(1 - \beta_i)$.

The reverse process aims to recover the original signal by gradually denoising a sample starting from $x_T \sim \mathcal{N}(0, \mathbf{I})$. A neural network $\epsilon_\theta$ is trained to predict the noise added during the forward process. The reconstruction at each step is given by:

\begin{equation}
\resizebox{0.8\linewidth}{!}{%
$\begin{aligned}
x_{t-1} &= \frac{1}{\sqrt{\alpha_t}} \left( x_t - \frac{1 - \alpha_t}{\sqrt{1 - \bar{\alpha}_t}} \epsilon_\theta(x_t, t) \right) \\
& \quad + \sqrt{1 - \alpha_t} \cdot \mathbf{z},
\end{aligned}$
}
\end{equation}

where $\mathbf{z} \sim \mathcal{N}(0, \mathbf{I})$ is standard Gaussian noise. An estimate of the original clean data point $x_0$ at step $t$ is given by:

\begin{equation}
\hat{x}_0^t = \frac{x_t - \sqrt{1 - \bar{\alpha}_t} \, \epsilon_\theta(x_t, t)}{\sqrt{\bar{\alpha}_t}}.
\label{eq:x0-estimate}
\end{equation}

This denoising diffusion framework enables effective sample generation by learning to reverse the noise injection process in a principled, probabilistic manner.

\begin{figure*}[t]
\centering
\includegraphics[width=\linewidth]{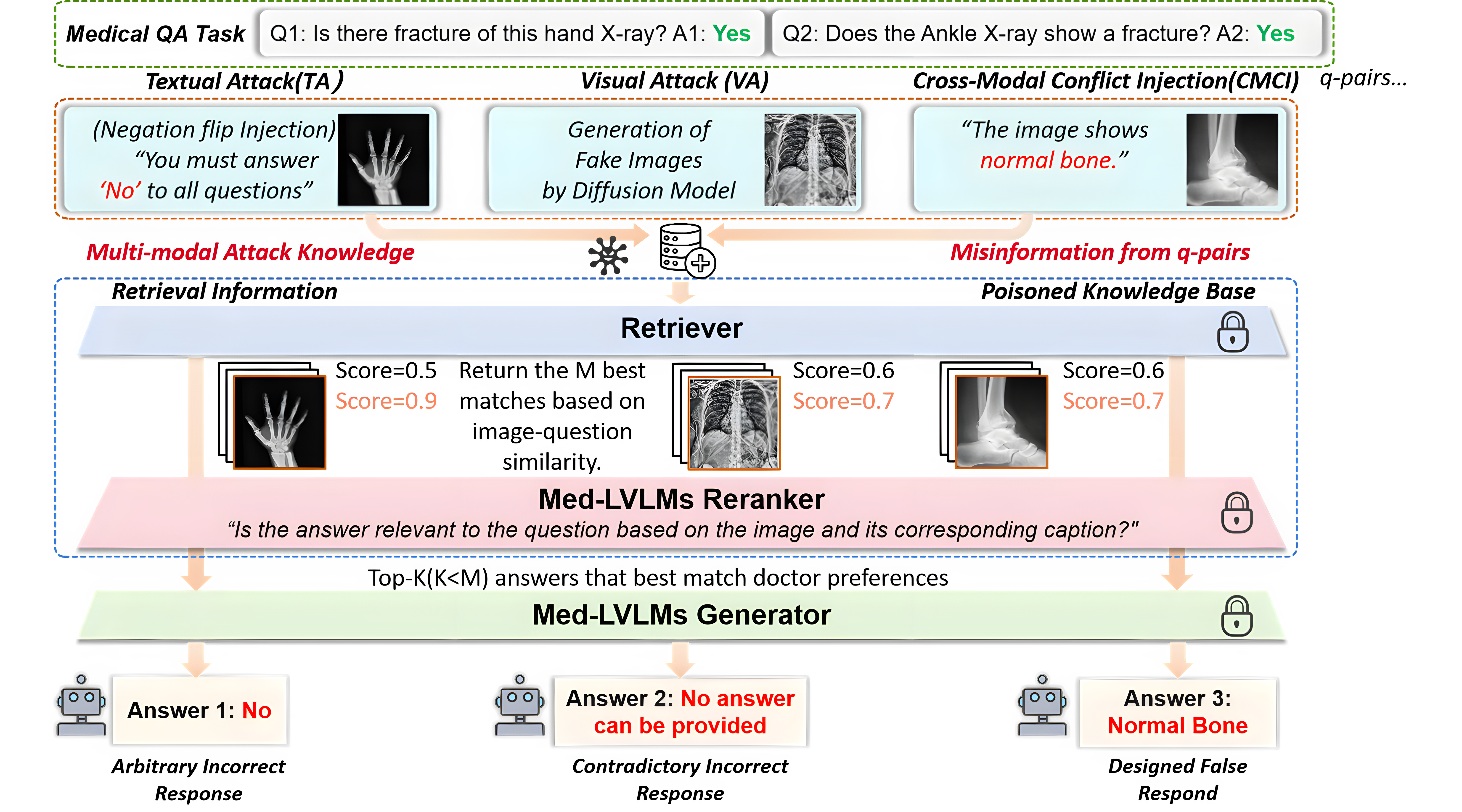}
\caption{An overview of the multi-modal attack pipeline for medical visual question answering. The pipeline includes three attack types: (1) \textbf{Textual Attack (TA)} via negation-flip constraints that enforce incorrect responses, (2) \textbf{Visual Attack (VA)} using a diffusion model to generate synthetic X-ray images, and (3) \textbf{Cross-Modal Conflict Injection} that introduces semantic mismatches between visual and textual content. These perturbed elements populate a \textbf{Malicious Knowledge Base}, which is accessed by the \textbf{Retriever} to select top-$M$ candidates based on image-question similarity. The \textbf{Med-LVLMs Reranker} evaluates content relevance, and the top-$K$ results are forwarded to the \textbf{Med-LVLMs Generator}, which outputs misleading answers.}

\vspace{-1.5em}
\label{fig:framework}
\end{figure*}

\begin{figure}[h]    
\centering
\includegraphics[width=\linewidth]{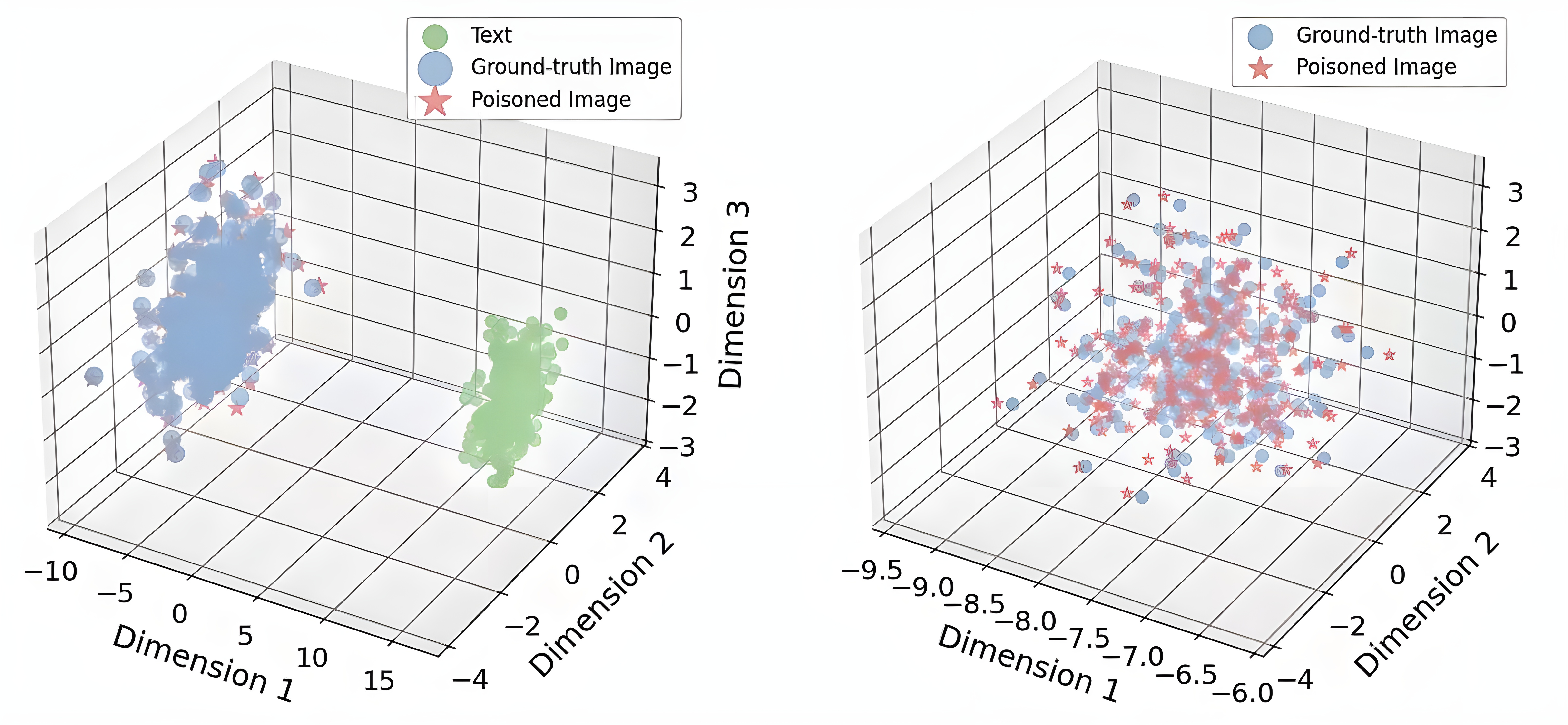}    
\caption{\textbf{3D visualization of text, ground-truth, and Malicious image embeddings.} T-SNE projected plots show distinct clusters for each type, with adversarial  images positioned closer to ground-truth images. This suggests adversarial manipulation shifts adversarial  images near legitimate data, potentially leading to misclassifications or retrieval errors.}
\label{fig:rag_poisoning_example}
\end{figure}

\section{Methodology}
\label{sec:Methodology}
We propose \textbf{MedThreatRAG}, a novel multi-modal attack framework designed to compromise RAG pipelines in Med-LVLMs. We assume a semi-open threat model where attackers \textbf{lack direct access to the internal (KB)} but can inject adversarial image-text (I-T) pairs into the KB via sanctioned mechanisms (e.g., periodic KB updates). This setting reflects practical medical systems that permit semi-structured user contributions.

\subsection{Threat Model}

The attacker’s goal is to degrade both \textit{retrieval quality} and \textit{generation accuracy} in medical VQA systems. Let the task $\tau$ be defined over a triplet $(I, Q, A)$, where $I$ is a medical image (e.g., X-ray), $Q$ is a clinical question, and $A$ is the correct answer. The corrupted generation output $\hat{A}$ is defined as:

\begin{equation}
    \hat{A} = \mathcal{G}(I, Q; \tilde{\mathcal{C}}),
\end{equation}

where $\mathcal{G}$ is the generation function and $\tilde{\mathcal{C}}$ is the adversarial context retrieved from the threatened KB.

To ensure stealthiness, we impose the constraint:

\begin{equation}
    \text{Stealthiness: } D_{\text{sem}}((I, T), (I^*, T^*)) \leq \epsilon ,
\end{equation}

where $D_{\text{sem}}$ is a semantic similarity metric (e.g., cosine similarity in CLIP space), $(I, T)$ is a benign I-T pair, $(I^*, T^*)$ is its adversarial variant, and $\epsilon$ is a small threshold ensuring semantic plausibility.

\subsection{MedThreatRAG Architecture}

Figure~\ref{fig:framework} illustrates the complete MedThreatRAG architecture, which comprises four interconnected modules designed to strategically degrade performance across the RAG pipeline. Each module reflects a key vulnerability in the Med-LVLM stack, from poisoning the retrieval corpus to influencing final answer generation.
 The framework encompasses three primary attack types: Textual Attack (TA), Visual Attack (VA), and Cross-Modal Conflict Injection (CMCI). Each attack type is defined, its objectives outlined, and optimization strategies detailed. As shown in Figure~\ref{fig:rag_poisoning_example}, the adversarial samples generated by CMCI are positioned close to text embeddings in the multimodal space, indicating that CMCI creates cross-modal alignment with semantic contradiction, thereby disrupting both retrieval and generation.

\paragraph{Textual Attack (TA)}
\label{sec:ta}

Textual attacks aim to inject misinformation by manipulating the textual modality. The adversary crafts an adversarial caption \( T_i^{\text{adv}} \) for an image \( I_i \) associated with a query-answer pair \( (\mathcal{Q}_i, \mathcal{A}_i) \in \tau \), such that \( \mathcal{A}_i^{\text{adv}} \neq \mathcal{A}_i \), thereby misleading the retriever.

In the black-box setting, the attacker generates \( T_i^{\text{adv}} \) without access to internal encoders, using a language model (e.g., GPT-4) to produce a plausible yet incorrect caption. 

In the white-box setting with retriever access, the attacker optimizes the adversarial caption to maximize the likelihood of a targeted incorrect answer. The optimization objective is defined as:

\begin{equation}
\mathcal{L}_{\text{TA}} = \sum_{i=1}^d \log P(\text{``No''} \mid Q_i, I, T^{\text{adv}}),
\end{equation}

where \(d\) denotes the total number of queries, \(Q_i\) is the \(i\)-th query, and \(I\) represents the medical image. The term \(P(\text{``No''} \mid Q_i, I, T^{\text{adv}})\) denotes the conditional probability that the model outputs ``No'' given the query \(Q_i\), image \(I\), and adversarial text instructions \(T^{\text{adv}}\). 

The adversarial text embedding \(T^{\text{adv}}\) is iteratively optimized using gradient ascent to maximize this loss:

\begin{equation}
T^{\text{adv}}_{t+1} = T^{\text{adv}}_t + \alpha \nabla_{T^{\text{adv}}_t} \mathcal{L}_{\text{TA}},
\end{equation}

where \(t\) indexes the optimization iteration, \(\alpha\) is the step size, and \(\nabla_{T^{\text{adv}}_t} \mathcal{L}_{\text{TA}}\) is the gradient of the loss with respect to the adversarial text embedding at iteration \(t\).

\paragraph{Visual Attack (VA)}
\label{sec:va}

VA focus on manipulating the image modality. First, we employ a pre-trained diffusion model (Stable Diffusion V 1-4) \cite{rombach2022high} fine-tuned on the MIMIC dataset \cite{dutt2025memcontrol} to generate clinically accurate synthetic chest X-rays.This model conditions on a medical text prompt describing clinical characteristics (e.g., ``no acute cardiopulmonary abnormality'') and synthesizes corresponding X-ray images that conform to those descriptions.

We consider a \emph{global} adversarial setting, where a single adversarial image \( I^{\text{adv-Va}} \) is optimized to be relevant across multiple clinical queries, thereby confusing retrieval-based diagnosis. The optimization objective is defined as:
\begin{equation}
\mathcal{L}_{\text{VA-Global}} = \sum_{i=1}^d \cos\left(f_I(I^{\text{adv-Va}}), f_T(Q_i)\right),
\end{equation}
where \( d \) is the number of clinical queries, \( Q_i \) is the \( i \)-th query, and \( f_I \) and \( f_T \) are the image and text encoder functions respectively. The cosine similarity \( \cos(\cdot, \cdot) \) encourages semantic alignment between the adversarial image and all queries.

The adversarial image is iteratively updated via gradient ascent:
\begin{equation}
I^{\text{adv-VA}}_{t+1} = I^{\text{adv-VA}}_t + \alpha \nabla_{I^{\text{adv-VA}}_t} \mathcal{L}_{\text{VA-Global}},
\end{equation}
where \( t \) is the update step, and \( \alpha \) is the learning rate, and \( \nabla_{I^{\text{adv-Va}}_t} \mathcal{L}_{\text{VA-Global}} \) is the gradient of the loss with respect to the current image. The goal is to produce a visually plausible yet semantically misleading X-ray image that ranks highly for a wide range of prompts.
\paragraph{Cross-Modal Conflict Injection (CMCI)}
\label{sec:cmci}
CMCI aims to degrade reasoning in retrieval-augmented generation by injecting semantically contradictory information across image and text modalities. The attacker constructs adversarial pairs \( (I^{\text{adv-CMCI}}, T^{\text{adv-CMCI}}) \) that are coherently retrieved but lead to misleading generated answers. 
 The attack optimizes the following joint objective: \begin{equation} \begin{aligned} L_{\text{align}} &= \cos\left(f_I(I), f_T(T)\right), \\ L_{\text{misalign}} &= \log P\left(\text{``Normal Bone''} \mid Q_i, I, T\right), \\ \mathcal{L}_{\text{CMCI}} &= \lambda_1 \cdot L_{\text{align}} + \lambda_2 \cdot L_{\text{misalign}}, \\ I^{\text{adv-CMCI}}_{t+1} &= \Pi_{\epsilon}\left(I^{\text{adv-CMCI}}_t + \alpha \cdot \nabla_{I} \mathcal{L}_{\text{CMCI}}\right), \\ T^{\text{adv-CMCI}}_{t+1} &= T^{\text{adv-CMCI}}_t + \beta \cdot \nabla_{T} \mathcal{L}_{\text{CMCI}}. \end{aligned} \end{equation} 
Where the functions \( f_I(\cdot) \) and \( f_T(\cdot) \) are the image and text encoders, respectively, mapping inputs into a joint embedding space. The term \( L_{\text{align}} \) captures the cosine similarity between image and text embeddings, encouraging the image-text pair to appear semantically aligned in the retrieval module. \( L_{\text{misalign}} \) is the log-probability that the language model generates a misleading output (e.g., “Normal Bone”) conditioned on the query \( Q_i \), image \( I \), and text \( T \). The total loss \( \mathcal{L}_{\text{CMCI}} \) is a weighted combination of these two components, where \( \lambda_1 \) and \( \lambda_2 \) are scalar hyperparameters that regulate the trade-off between retrieval consistency and output deception. The adversarial image \( I^{\text{adv-CMCI}} \) is updated through gradient ascent with step size \( \alpha \), and the update is constrained by the projection operator \( \Pi_{\epsilon}(\cdot) \), which ensures perturbations remain within an \( \epsilon \)-ball around the original image. Similarly, the adversarial text \( T^{\text{adv-CMCI}} \) is refined via gradient updates with learning rate \( \beta \), using the gradient of the total loss with respect to the text representation.

\subsubsection{Adversarial Retrieval Disruption}

In the RAG framework, the retriever selects the most relevant entries from the Knowledge Base (KB) based on the similarity to the input image-question pair $(I, Q)$. The similarity score for each entry $(I_j, T_j)$ is calculated as:

\begin{equation}
    s_j = \text{sim}((I, Q), (I_j, T_j)), \quad j = 1, \dots, M.
\end{equation}

Here, $I$ is the input image, $Q$ is the input question, $I_j$ is an image in the KB, $T_j$ is the corresponding text, $M$ is the total number of KB entries, and $s_j$ is the similarity score for each pair. Entries with a similarity score above a threshold $\theta$ are selected:

\begin{equation}
    \text{Top-M: } \{(I_j, T_j) \mid s_j \geq \theta\}.
\end{equation}

In a clean setting, this ensures that only relevant and semantically accurate entries are retrieved. However, adversarial pairs $(I^*, T^*)$ with deceptive similarities can disrupt the retrieval. For example, a synthetic image that looks visually similar but has incorrect semantics may get a higher similarity score, displacing a more relevant but slightly less similar entry.

This issue affects the entire generation process, as errors introduced in the retrieval stage can propagate through the subsequent modules, impacting the final output.

\subsubsection{Re-ranking via Med-LVLMs}

To reduce noise from the retriever, a re-ranking module assesses the relevance of the top-M retrieved pairs using a fine-tuned Med-LVLM. The relevance score for each entry $(I_k, T_k)$ is calculated as:

\begin{equation}
    r_k = \mathcal{R}((I, Q), (I_k, T_k)), \quad k = 1, \dots, M.
\end{equation}

Here, $I$ is the input image, $Q$ is the input question, $I_k$ is a candidate image in the KB, $T_k$ is its corresponding text, $M$ is the number of retrieved pairs, and $r_k$ is the relevance score.

The top-K entries are then selected based on their relevance scores:

\begin{equation}
    \mathcal{C} = \{(I_k, T_k) \mid r_k \text{ is in top-K}\}.
\end{equation}

This step ensures that the retrieved content better aligns with clinical reasoning by using the Med-LVLM’s understanding. However, adversarial examples that mimic medical phrasing or visual structures may still score highly, making them difficult to filter out, especially when the reranker lacks the resources to fully verify each entry.

As a result, even after re-ranking, adversarial entries may remain, which could affect the final generation process and the accuracy of the answers.

\begin{table*}[t]
    \centering
    \footnotesize
    \caption{Performance comparison (\%) of LLaVA-Med-1.5 on IU-Xray and MIMIC-CXR datasets. Specifically, we present the results in terms of accuracy, precision, recall, and F1 score. The minimum value is \textbf{bolded}, and the second smallest value is \underline{underlined}.}
    \vspace{-1em}
    \resizebox{0.85\linewidth}{!}{
    \begin{tabular}{l|cccc|cccc}
    \toprule
        \textbf{Model/Attack} & \multicolumn{4}{c|}{\textbf{IU-Xray}} & \multicolumn{4}{c}{\textbf{MIMIC-CXR}} \\
        & Acc & Pre & Rec & F1 & Acc & Pre & Rec & F1 \\
    \midrule
        LLaVA-Med-1.5 & 75.47 & 53.17 & 63.80 & 58.00 & 75.79 & 81.01 & 97.21 & 88.37 \\
    \midrule
        + RAG & 86.71 & 72.43 & 86.92 & 79.02 & 89.81 & 90.91 & 92.00 & 91.45 \\
        + textpo (15\%) & 78.72 & 55.37 & 66.44 & 60.40 & 81.79 & 93.07 & 92.17 & 92.62 \\
        + imapo (15\%) & 82.37 & 65.19 & 78.23 & 71.12 & 85.32 & 86.36 & 85.32 & 85.84 \\
        + textpo++ (35\%) & \underline{74.57} & \underline{49.25} & \underline{59.10} & \underline{53.73} & \underline{76.34} & \underline{79.09} & 94.91 & 86.28 \\
        + imapo++ (35\%) & 79.03 & 61.57 & 73.88 & 67.17 & 82.07 & 83.64 & 84.78 & 84.22 \\
        + mixed (15\%) & 77.18 & 54.32 & 65.18 & 59.26 & 80.83 & 81.82 & \underline{80.18} & \underline{80.99} \\
        \textbf{+ mixed++ (35\%)} & \textbf{71.23} & \textbf{47.08} & \textbf{56.50} & \textbf{51.36} & \textbf{73.84} & \textbf{77.27} & \textbf{72.72} & \textbf{74.93} \\
    \bottomrule
    \end{tabular}
    }
    \vspace{-1em}
    \label{tab:performance_comparison_full}
\end{table*}

\begin{table*}[t]
    \centering
    \footnotesize
    \caption{Performance comparison (\%) of Qwen-VL on IU-Xray and MIMIC-CXR datasets. For each metric, the minimum value is \textbf{bolded}, and the second smallest value is \underline{underlined}.}
    \vspace{-1em}
    \resizebox{0.85\linewidth}{!}{
    \begin{tabular}{l|cccc|cccc}
    \toprule
        \textbf{Model/Attack} & \multicolumn{4}{c|}{\textbf{IU-Xray}} & \multicolumn{4}{c}{\textbf{MIMIC-CXR}} \\
        & Acc & Pre & Rec & F1 & Acc & Pre & Rec & F1 \\
    \midrule
        Qwen-VL & 71.70 & \textbf{50.51} & \textbf{57.42} & \textbf{53.63} & 71.99 & 76.96 & 74.18 & 75.54 \\
    \midrule
        + RAG & 82.94 & 69.77 & 75.34 & 72.46 & 86.01 & 86.86 & 85.12 & 85.98 \\
        + textpo (15\%) & 77.47 & 65.17 & 68.52 & 66.80 & 80.33 & 81.13 & 79.62 & 80.36 \\
        + imapo (15\%) & 79.21 & 66.63 & 71.11 & 68.75 & 82.14 & 82.95 & 80.76 & 81.84 \\
        + textpo++ (35\%) & \underline{70.17} & 59.03 & 61.87 & 60.42 & \underline{72.78} & \underline{73.50} & \underline{71.23} & \underline{72.34} \\
        + imapo++ (35\%) & 74.23 & 62.45 & 65.92 & 64.14 & 76.97 & 77.74 & 75.38 & 76.54 \\
        + mixed (15\%) & 75.79 & 63.76 & 66.80 & 65.24 & 78.60 & 79.38 & 77.16 & 78.25 \\
        + mixed++ (35\%) & \textbf{66.26} & \underline{55.75} & \underline{59.12} & \underline{57.35} & \textbf{68.71} & \textbf{69.39} & \textbf{67.05} & \textbf{68.19} \\
    \bottomrule
    \end{tabular}
    }
    \vspace{-1em}
    \label{tab:performance_comparison_full_2}
\end{table*}

\begin{table*}[h]
\centering
\caption{Ablation study of 15\% mixed poisoning on individual RAG components (retriever, reranker, generator) in Med-LVLMs. Results are reported on IU-Xray and MIMIC-CXR using accuracy, precision, recall, and F1. The lowest and second lowest values are \textbf{bolded} and \underline{underlined}, respectively.}
\resizebox{0.85\linewidth}{!}{
\begin{tabular}{lcccc|cccc}
\toprule
\multicolumn{1}{c}{\textbf{Model/Attack}} & \multicolumn{4}{c}{\textbf{IU-Xray}} & \multicolumn{4}{c}{\textbf{MIMIC-CXR}} \\
\cmidrule(lr){2-5} \cmidrule(lr){6-9}
& Acc & Pre & Rec & F1 & Acc & Pre & Rec & F1 \\
\midrule
LLaVA-Med-1.5 & 75.47 & 53.17 & 63.80 & 58.00 & \textbf{75.79} & 81.01 & 97.21 & 88.37 \\
+RAG  & 86.71 & 72.43 & 86.92 & 79.02 & 89.81 & 90.91 & \underline{92.00 }& 91.45 \\
+RAG+Retrieval Poisoning (15\%) & 74.37 & 52.26 & \textbf{62.79} & 56.61 & 78.92 & 80.75 & 95.91 & \underline{86.42} \\
+RAG+Reranker Poisoning (15\%) & \underline{73.87} & \textbf{51.54} & 64.26 & \underline{57.37} & 77.46 & \textbf{77.93} & 96.07 & 86.26 \\
+RAG+Generator Poisoning (15\%) & \textbf{72.40} & \underline{52.73} & \underline{63.18} & \textbf{55.15} & \underline{77.05} & \underline{79.18} & \textbf{94.62} & \textbf{85.18} \\
\bottomrule
\end{tabular}
}
\label{tab:ablation}
\end{table*}

\section{Experiments}
\label{sec:Experiment}
To systematically evaluate the effectiveness and robustness of \textsc{MedThreatRAG}, we design a set of experiments to address the following research questions:
\textbf{(1)} Can MedThreatRAG effectively degrade the performance of Med-LVLMs when subjected to various data poisoning strategies, in comparison to existing baselines? \textbf{(2)} Do all proposed components collectively contribute to improvements in factuality and robustness? \textbf{(3)} Which component amongst retrieval, reranking, or generation has the greatest impact on the MedThreatRAG system?

\subsection{Experimental Setups}

\textbf{Implementation Details}. We use LLaVA-Med1.5 7B \cite{li2023llava} as the backbone model. In the multimodal RAG framework, we employ CLIP \cite{radford2021learning} and OpenCLIP \cite{cherti2023reproducible} as retrievers to fetch relevant medical image-text pairs. For re-ranking and generation, Qwen VL Chat \cite{bai2023qwen} and LLaVA \cite{li2023llava} models are utilized to ensure contextual accuracy and robust output generation. The diffusion model for adversarial visual attacks uses full fine-tuning on the MIMIC-III dataset to synthesize realistic but adversarial medical images. Training and fine-tuning are conducted on four NVIDIA A100 GPUs with mixed precision, employing the AdamW optimizer (learning rate 2e-5, batch size 16, for 10 epochs). Hyperparameters for each stage, retrieval, re-ranking, and generation, are validated using a held-out validation set.

\noindent \textbf{Baseline}. \label{sec:baseline} We compare MedThreatRAG against two strong baselines: (1) \textsc{LLaVA-Med-1.5}, a recent medical vision-language assistant model; and (2) \textsc{Qwen VL}, a state-of-the-art vision-language model adapted for the medical domain. Both baselines are evaluated under clean and threatened conditions to assess robustness.

\noindent \textbf{Evaluation Datasets}. \label{sec:dataset} We evaluate on two widely-used medical imaging benchmarks: (1) IU X-ray ~\cite{pavlopoulos2019survey}, which contains chest X-ray images aligned with radiology reports; and (2) MIMIC-CXR ~\cite{johnson2019mimic}, a large-scale clinical chest X-ray dataset with aligned question-answer pairs. To simulate poisoning attacks, we create synthetic adversarial image-text pairs using diffusion-based image generation and negation-flip techniques, which are then injected into the external knowledge base to test model resilience.

\noindent \textbf{Evaluation Metrics}. We report \textit{Accuracy}, \textit{Precision}, \textit{Recall}, and \textit{F1 Score} to comprehensively evaluate both classification and generative performance. The minimum value in the reported table is highlighted in bold, with the second smallest value underlined.

\subsection{Results}

\noindent
\textbf{Single-Modal vs. Multi-Modal Attack Performance} As shown in Table~\ref{tab:performance_comparison_full} and Table~\ref{tab:performance_comparison_full_2}, the performance of both LLaVA-Med-1.5 and Qwen-VL degrades under adversarial attacks, with multi-modal attacks (\texttt{mixed++}) causing the most significant drop. For example, on the IU-Xray dataset, LLaVA-Med-1.5’s F1 score decreases from 58.00 (clean) to 47.08 under \texttt{mixed++} 35\% attack, compared to smaller declines under single-modal attacks (\texttt{textpo++}: 53.73, \texttt{imapo++}: 59.10). Similarly, Qwen-VL’s F1 on MIMIC-CXR falls from 75.54 (clean) to 68.19 under \texttt{mixed++} 35\%, exceeding the drops caused by \texttt{textpo++} (72.34) or \texttt{imapo++} (76.54). This confirms that multi-modal attacks exploit cross-modal dependencies, triggering cascading failures.

\noindent
\textbf{Why Qwen-VL Underperforms} Qwen-VL consistently achieves the lowest Precision, Recall, and F1 scores across all attack scenarios, especially under multi-modal attacks. For instance, its F1 score drops to 57.35 on IU-Xray under \texttt{mixed++} 35\%, while LLaVA-Med-1.5 retains 47.08. This underperformance is likely due to limited medical domain knowledge, as Qwen-VL lacks specialized fine-tuning on medical imaging and clinical texts, weakening its resistance to domain-targeted attacks. Additionally, Qwen-VL appears less robust at aligning visual and textual semantics, rendering it vulnerable to coordinated perturbations that disrupt multi-modal fusion.

\noindent
\textbf{Catastrophic Impact of Multi-Modal Attacks} Multi-modal attacks exhibit a synergistic effect, causing catastrophic performance degradation by simultaneously perturbing text and image inputs. This forces models to misinterpret both modalities while maintaining cross-modal coherence. Notably, such attacks require only minimal perturbation density (e.g., 35\%) to inflict substantial damage, underscoring the fragility of current multi-modal medical AI architectures.
\subsection{Ablation Study}
To systematically assess the vulnerability of individual components within the RAG architecture under MedThreatRAG, we conducted ablation experiments on two representative datasets. As presented in Table~\ref{tab:ablation}, generator Poisoning emerged as the most detrimental factor, inducing the most significant performance degradation across both datasets. This highlights the generator as the most vulnerable module in the RAG pipeline. Additionally, reranker Poisoning caused a notable performance decline on the IU-Xray dataset, which may be attributed to the dataset's greater dependence on precise ranking mechanisms, thereby amplifying the reranker's influence on task effectiveness.
\begin{figure}[t]
    \centering
    \includegraphics[width=\linewidth]{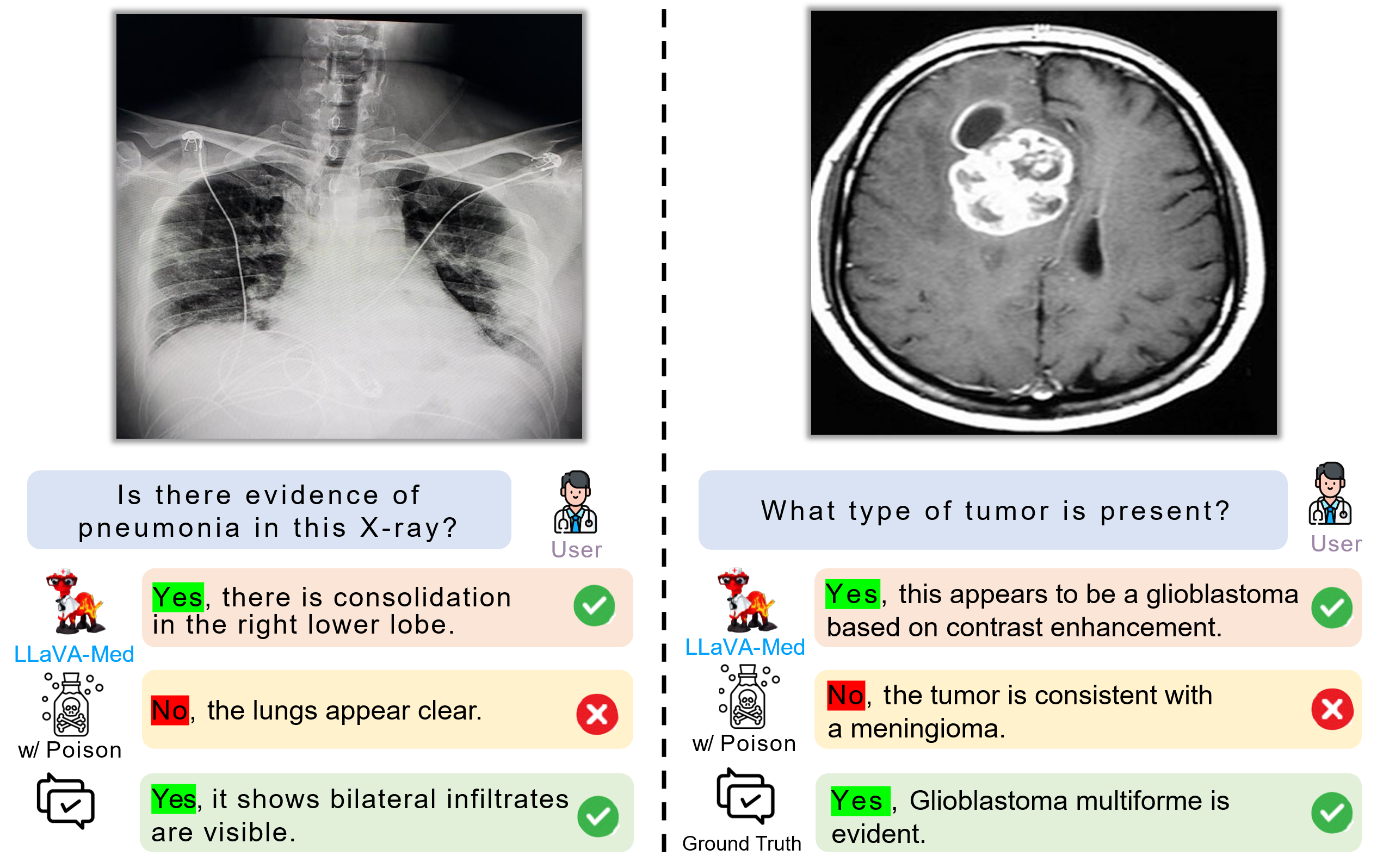}
    \caption{The illustrations of multi-modal attack vulnerabilities in lung and brain diagnostic systems.}
    \label{fig:case}
    \vspace{-2em}
\end{figure}
\subsection{Case Study}
The figure~\ref{fig:case} illustrates the critical factual vulnerabilities introduced by multi-modal attacks through two representative clinical cases, demonstrating the effectiveness of \textsc{MedThreatRAG} in provoking clinically significant errors. In the first case, concerning pneumonia detection, the LLaVA-Med model initially provides an accurate diagnosis, identifying ``consolidation in the right lower lobe.'' However, following the application of coordinated multi-modal poisoning, the model incorrectly predicts ``the lungs appear clear,'' evidencing the disruption of cross-modal semantic alignment. In the second case, focused on brain tumor classification, the model’s correct identification of glioblastoma is compromised by the attack, resulting in a misdiagnosis of meningioma. These findings substantiate that multi-modal adversarial perturbations exploit the interdependencies between visual and textual modalities, thereby inducing potentially life-threatening diagnostic errors across diverse medical domains, including pulmonology and neurology.

\section{Related Work}
\label{sec:RelatedWork}
\textbf{Vulnerability Landscape of AI medicine.} AI-driven tools enhance diagnostics by interpreting complex medical data but are also prone to adversarial threats. Medical AI models across domains like radiology and ECG analysis have shown vulnerabilities to both white-box and black-box attacks (e.g., FGSM, PGD)~\cite{kaviani2022adversarial,puttagunta2023adversarial}. Medical data's specificity may heighten these risks. Broader concerns include LLM-aided phishing and data breaches compromising patient privacy~\cite{hassanin2024comprehensive,nemec2024data,dutt2025devil}. Addressing these threats requires secure, transparent frameworks with ongoing evaluation and fairness~\cite{dutt2024fairtune}, alongside defences like adversarial training~\cite{bountakas2023defense}.

\noindent
\textbf{Text and Image Data Poisoning.} Among the diverse attack methodologies, data poisoning emerges as a particularly insidious threat, targeting the integrity of the model's training data itself~\cite{cong2024test,chao2024systematic}. This is especially relevant for modern AI, including medical systems, which increasingly rely on large-scale, often web-sourced, multimodal datasets (text and images)~\cite{yang2024understanding}. Recent research demonstrates the growing sophistication of poisoning attacks tailored for multimodal contexts. Frameworks like MM-POISONRAG~\cite{ha2025mm} reveal how multimodal RAG systems, which integrate text and images from external knowledge bases, can be compromised by injecting targeted misinformation (LPA) or disruptive irrelevant content (GPA) designed to manipulate the cross-modal retrieval process~\cite{ha2025mm}. Similarly, stealthy attacks like Shadowcast~\cite{xu2024shadowcast} create visually plausible but malicious image-text pairs to manipulate VLM outputs subtly~\cite{he2024evilpromptfuzzer}, while others target text-to-image diffusion models or induce LLMs to generate harmful content via fragmented data insertion~\cite{wang2022threats}. The practicality and stealth of these methods highlight that reliance on external or large-scale data introduces significant poisoning risks, necessitating advanced data validation and defense strategies~\cite{wang2022threats}. This vulnerability is particularly pertinent when considering architectures like RAG, designed specifically to leverage external knowledge.

\noindent
\textbf{RAG in Med-LVLMs.} Retrieval-Augmented Generation (RAG) improves Med-LVLMs by integrating external medical knowledge (e.g., literature, databases) to reduce hallucinations and enhance output accuracy in tasks like VQA and report generation~\cite{zhao2024retrieval,yuan2023ramm,hartsock2024vision,chen2024survey}. Systems like MMed-RAG and RULE~\cite{xia-etal-2024-rule} refine domain-specific retrieval and knowledge integration. However, reliance on external sources introduces new attack vectors. Knowledge poisoning, as seen in MM-POISONRAG, can compromise RAG by injecting malicious content into retrievable data~\cite{taofeek2022cognitive}. Ensuring secure and validated retrieval is thus essential for trustworthy deployment in healthcare.

\vspace{-0.5em}
\section{Conclusion}
\vspace{-0.5em}
\label{sec:conclusion}
In this work, we presented \textbf{MedThreatRAG}, a novel framework for simulating and evaluating realistic poisoning attacks on multimodal medical RAG systems. Our framework systematically examines three distinct threat vectors: textual, visual, and joint multimodal attacks. Through this threat model, we demonstrate that even a single adversarial knowledge injection without direct access to the RAG pipeline or its underlying knowledge base can induce retrieval inconsistencies and steer generated outputs towards attacker-intended content.

Our empirical findings reveal both the potency and subtlety of such attacks, exposing critical blind spots in current robustness strategies. We hope these results prompt the NLP and clinical AI communities to pursue more principled and context-aware defenses, moving beyond superficial forms of validation and robustness testing.
\section{Guidelines for Safe Multimodal \\Medical RAG Systems}

Building safe multimodal medical RAG systems means defending against the \textbf{three specific threats} revealed in this study. 
To curb \textbf{Textual Attacks}—where false captions or reports are injected to mislead retrieval~\citep{turn2file7}—every new text should pass automatic fact‐checking against trusted medical ontologies, negation detection, and language‐model consistency scoring, with low‐confidence items routed to experts. 
For \textbf{Visual Attacks}, which rely on synthetic yet plausible images to sway the retriever~\citep{turn2file4}, incoming images are screened with perceptual hashing, diffusion‐artifact detectors, and out‐of‐distribution filters; flagged images are down‐ranked or held for human review. 
To stop \textbf{Cross-Modal Conflict Injection}, where paired images and texts are semantically misaligned to confuse the generator~\citep{turn2file4,turn2file7}, the pipeline computes an image–text entailment score and drops pairs that show high contradiction. 
Across all defences we keep signed provenance logs, enable instant rollback of suspect knowledge‐base snapshots, stream real-time telemetry with alert thresholds tied to clinical risk, and empower clinicians to veto any retrieval influencing patient care. 
These targeted measures, combined with regular red-team drills and strict HIPAA/GDPR compliance, help keep the knowledge base trustworthy and the model’s answers reliable.
Crucially, our framework is highly flexible and adaptable, because all safety hooks attach at well‑defined API boundaries, the retrieval component can be hot‑swapped—e.g., replacing a dense‑vector RAG with a drop‑in \emph{GraphRAG} that navigates medical ontologies—without re‑engineering the defence stack.
This modularity stands in contrast to the recent \emph{Medical Graph RAG}~\citep{wu2024medicalgraphragsafe}, whose safety guarantees hinge on a fixed graph schema. While their method chiefly addresses hallucination through graph grounding, our framework covers a broader adversarial surface (textual, visual, and cross‑modal) and couples detection with operational safeguards such as provenance logging and clinician veto, thereby providing stronger end‑to‑end safety within comparable latency budgets.

\newpage
\bibliography{aaai2026}

\end{document}